\documentclass[11pt]{article}
\usepackage[preprint]{acl}
\usepackage{times}
\usepackage{latexsym}
\usepackage[T1]{fontenc}
\usepackage[utf8]{inputenc}
\usepackage{microtype}
\usepackage{graphicx}
\usepackage{booktabs}
\usepackage{fancyvrb}
\usepackage{url}

\title{PaliBench: A Multi-Reference Blueprint for Classical Language Translation Benchmarks}

\author{
  M\'at\'e Metzger \\
  Independent Researcher \\
  \texttt{metzgermate@gmail.com}
  \And
  Nadnapang Phophichit \\
  International Buddhist Studies College \\
  Mahachulalongkornrajavidyalaya University \\
  \texttt{nadnapang@ibsc.mcu.ac.th}
}

\begin{document}
\maketitle
\begin{abstract}
Digital humanities projects increasingly rely on machine translation and large language models to widen access to classical, religious, and otherwise under-translated textual traditions. Yet standard translation benchmarks are poorly suited to such materials: they typically compare a system output against a single reference translation, even though classical texts often support multiple faithful renderings that differ in terminology, register, and interpretation. This article introduces PaliBench, both a benchmark for Pali-to-English translation and a reusable method for constructing multi-reference translation benchmarks for classical languages. The Pali case study draws on passages from the Sutta Pitaka aligned with independent English translations by Bhikkhu Sujato, Bhikkhu Thanissaro, and Bhikkhu Bodhi. The workflow combines LLM-assisted alignment of independently segmented translations, automated verification against source files, passage-level quality filtering, deduplication of formulaic repetitions, and multi-metric evaluation against multiple human references. The resulting benchmark contains 1,700 passages spanning 8,389 segments and approximately 345,000 tokens. We use it to evaluate ten contemporary large language models with complementary metrics, finding strong cross-metric concordance in system rankings alongside substantial variation in reliability and semantic outlier rates. The broader contribution is methodological: PaliBench shows how existing scholarly translations can be transformed into evaluation infrastructure for interpretive textual traditions without treating any single translation as definitive. Although developed for Pali Buddhist texts, the approach could be portable to other classical corpora where sufficient independent reference translations exist.
\end{abstract}


\section{Introduction}

Classical and religious text traditions present a recurring problem for digital humanities research: the available corpus is often large, culturally significant, and only partially translated, while the expertise required for translation is scarce. Machine translation and large language models therefore offer an attractive means of widening access, but evaluating their output is difficult. Standard machine translation benchmarks often compare a system output against a single reference translation. That assumption is poorly matched to classical texts, where translation is often interpretive and where multiple established translations may disagree substantially while remaining faithful to the source.

The Pali Tipitaka, the canonical scripture of Theravada Buddhism, offers a particularly clear case. Since the pioneering work of the Pali Text Society in the late nineteenth century, generations of scholars have produced English translations of its discourses, monastic rules, and philosophical treatises. These translations differ not only in era and scholarly convention but in fundamental interpretive commitments: where one translator renders a key doctrinal term with a Latinate philosophical equivalent, another may adopt a colloquial English expression, and a third may retain the Pali term with contextual glossing. All three approaches can be equally faithful to the source. This irreducible plurality of valid translation is not a deficiency to be resolved but a feature intrinsic to the translation of religious and philosophical material.

Machine translation has advanced rapidly in recent years, driven by the Transformer architecture \citep{vaswani-etal-2017-attention} and the scaling of large language models. Systems such as GPT-4, Claude, and dedicated multilingual models like NLLB \citep{nllb-etal-2022-no} now produce fluent output across hundreds of languages. Yet their performance on classical and philosophical languages remains poorly understood. Pali occupies an especially underserved position: it is not included in major multilingual MT benchmarks such as FLORES \citep{goyal-etal-2022-flores}, it likely has minimal representation in the pretraining data of most LLMs, and no multi-reference benchmark exists for assessing Pali-to-English translation quality. Researchers and practitioners working at the intersection of Buddhist studies and computational linguistics thus lack the basic infrastructure needed to measure progress.

Compounding this gap is a methodological problem. Standard MT evaluation relies on comparing system output against a single reference translation, using metrics ranging from lexical overlap measures such as BLEU \citep{papineni-etal-2002-bleu} and chrF \citep{popovic-2015-chrf} to neural learned metrics such as COMET \citep{rei-etal-2020-comet} and BLEURT \citep{sellam-etal-2020-bleurt}. The WMT22 Metrics Shared Task \citep{freitag-etal-2022-results} has demonstrated that neural metrics correlate far more strongly with expert human judgments than their lexical predecessors. However, all reference-based metrics share a structural limitation: they reward proximity to a specific rendering and penalize departures from it, even when those departures represent equally valid translations. For Pali Buddhist texts, where the space of legitimate translations is wide and the interpretive stakes are high, single-reference evaluation is not merely imprecise---it is systematically biased toward whichever scholarly tradition produced the reference.

Recent work on multi-reference evaluation has begun to address this limitation. \citet{fomicheva-etal-2020-multi} showed that incorporating translation variability estimates improves metric--human correlation, and \citet{wu-etal-2024-multiple} demonstrated that multiple references with meaningful paraphrastic variation yield significant gains in automatic metric reliability for literary translation. In the domain of classical language NLP, the MITRA project (\citealp{nehrdich-keutzer-2026-mitra}) addresses Buddhist canonical languages including Pali in its broader framework, while its large-scale parallel corpus is centered on Sanskrit, Chinese, and Tibetan sentence pairs. PaliBench differs by constructing a multi-reference Pali-to-English benchmark from independent published English translations.

This paper introduces PaliBench as both a concrete benchmark for Pali-to-English machine translation and a methodological blueprint for constructing multi-reference translation benchmarks in other classical language contexts. PaliBench comprises carefully selected passages from the Sutta Pitaka, each paired with multiple independent reference translations drawn from established scholarly sources. We evaluate ten contemporary LLMs on PaliBench using a complementary suite of lexical, neural, semantic, and diagnostic metrics. Our contributions are threefold. First, we provide, to our knowledge, the first multi-reference benchmark for Pali-to-English MT, enabling reproducible comparison of translation systems on Buddhist canonical texts. Second, we demonstrate a portable workflow for transforming independently segmented scholarly translations into benchmark infrastructure through alignment, verification, filtering, deduplication, and multi-reference evaluation. Third, we present empirical evidence on how model rankings and failure modes behave under multiple divergent references. The broader claim is that this workflow can be adapted to any classical language corpus with enough independent reference translations to support statistically meaningful evaluation.

\section{Literature Review}

\subsection{Automatic Evaluation of Machine Translation}

The evaluation of machine translation has undergone a fundamental transformation over the past two decades. The foundational Transformer architecture \citep{vaswani-etal-2017-attention} enabled the rapid advancement of neural MT systems, which in turn demanded evaluation methods capable of capturing semantic adequacy beyond surface-level overlap. Early automatic metrics centered on n-gram precision: BLEU \citep{papineni-etal-2002-bleu} introduced modified n-gram precision with a brevity penalty that became the de facto standard, while METEOR \citep{lavie-denkowski-2009-meteor} incorporated recall, stemming, and synonym matching to better accommodate legitimate translation variability. Character-level metrics such as chrF \citep{popovic-2015-chrf} offered robustness for morphologically rich languages by operating below the word level, and SacreBLEU \citep{post-2018-call} addressed reproducibility concerns by standardizing tokenization and reference handling. However, these overlap-based metrics correlate poorly with expert human judgments of modern high-quality MT systems. The WMT22 Metrics Shared Task \citep{freitag-etal-2022-results} demonstrated definitively that neural learned metrics---particularly COMET \citep{rei-etal-2020-comet} and BLEURT \citep{sellam-etal-2020-bleurt}---significantly outperform lexical metrics across domains and challenge sets when evaluated against Multidimensional Quality Metrics (MQM) annotations. COMET, a cross-lingual framework trained on human quality judgments using multilingual embeddings, is now a widely used neural metric for MT evaluation. Its extension, xCOMET \citep{guerreiro-etal-2024-xcomet}, adds interpretability through fine-grained error detection and achieves state-of-the-art correlation with MQM scores. BERTScore \citep{zhang-etal-2020-bertscore} offered an earlier embedding-based approach, computing token-level cosine similarity over BERT contextual representations. More recently, LLM-as-judge approaches such as GEMBA-MQM \citep{kocmi-federmann-2023-gemba} have shown that GPT-based quality estimation can achieve strong system-level ranking performance, raising questions about the future role of reference translations in evaluation. \citet{thompson-post-2020-paraphrasing} further demonstrated that framing evaluation as paraphrastic scoring---where a multilingual paraphraser scores MT output conditioned on a reference---outperforms prior metrics across 39 languages without requiring human judgments for training.

These developments establish a clear trajectory from surface matching toward semantic adequacy measurement, yet they also reveal a persistent limitation: most metrics and evaluation benchmarks have been developed for and validated on high-resource language pairs with abundant parallel data. The adequacy of these metrics for classical, liturgical, or otherwise specialized language domains remains largely untested.

\subsection{Large Language Models for Machine Translation}

The emergence of large language models as translation engines has reshaped the MT landscape. \citet{hendy-etal-2023-gpt} provided a comprehensive evaluation of GPT models across 18 translation directions, finding them competitive on high-resource pairs but significantly weaker on low-resource languages---a pattern with direct implications for ancient language translation. \citet{zhu-etal-2024-multilingual} extended this analysis to eight LLMs across massive multilingual settings, revealing that GPT-4 surpasses the supervised NLLB baseline in roughly 41\% of translation directions while still lagging behind commercial systems on low-resource languages. Their finding that cross-lingual in-context exemplars outperform same-language-pair exemplars for low-resource translation suggests promising strategies for languages like Pali, where parallel data is scarce but related languages may offer transferable signal. Dedicated MT fine-tuning has further closed this gap. \citet{xu-etal-2024-paradigm} introduced ALMA, a two-stage approach---monolingual pretraining followed by fine-tuning on small high-quality parallel data---that enables 7B-parameter LLMs to match GPT-3.5 and NLLB-54B. \citet{jiao-etal-2023-chatgpt} provided complementary evidence through their systematic evaluation of ChatGPT as a translator, documenting its facility with high-resource European languages and its struggles with distant and low-resource pairs. Together, these studies establish the experimental template that PaliBench extends to classical Buddhist texts: multi-system, multi-metric evaluation with attention to the specific failure modes that emerge when LLMs encounter languages at the margins of their training distributions.

\subsection{Low-Resource and Multilingual Translation Benchmarks}

The challenge of evaluating MT quality for low-resource languages has spurred dedicated benchmark construction. The FLORES-101 benchmark \citep{goyal-etal-2022-flores} established a controlled evaluation framework comprising 3,001 professionally translated sentences across 101 languages, enabling consistent cross-lingual comparison with standardized quality controls. The No Language Left Behind initiative \citep{nllb-etal-2022-no,nllb-team-2024-scaling} dramatically expanded multilingual MT to 200 languages using Sparsely Gated Mixture-of-Experts architectures, achieving a 44\% BLEU improvement over prior baselines. NLLB's methodology (cross-lingual transfer learning combined with large-scale evaluation via FLORES-200) demonstrates the viability of extending MT to severely underresourced languages, though its coverage does not include classical or liturgical languages such as Pali. PaliBench shares methodological DNA with these benchmarks: professionally curated reference translations, controlled quality, and multi-metric evaluation. However, it diverges in a critical respect. Where FLORES and NLLB evaluate breadth across living languages, PaliBench evaluates depth within a single classical language through multiple independent reference translations, foregrounding the problem of legitimate translation variability that is especially acute in religious and philosophical texts.

\subsection{NLP for Classical and Ancient Languages}

Computational approaches to ancient and classical languages have matured considerably in recent years, though they remain far behind the state of the art for modern languages. \citet{sommerschield-etal-2023-survey} survey the field comprehensively, documenting progress in ancient text processing across tasks from optical character recognition to machine translation while identifying persistent bottlenecks in data scarcity and morphological complexity. \citet{assael-etal-2022-restoring} demonstrated the power of human--AI collaboration for ancient Greek inscription restoration with Ithaca, showing that historian performance on restoration improved from 25\% to 72\% when assisted by the model---an instructive example of how computational tools can augment rather than replace domain expertise. For the Indo-Aryan language family most relevant to Pali, substantial progress has been made on Sanskrit. \citet{hellwig-2015-morphological} addressed the acute challenge of morphological disambiguation in classical Sanskrit, where fusional morphology, sandhi, and compound formation create extreme ambiguity that complicates downstream NLP tasks. \citet{bamman-burns-2020-latin} established the paradigm of domain-specific pretrained language models for classical languages with Latin BERT, trained on 642.7 million words spanning the Classical era to the 21st century, achieving state-of-the-art performance on POS tagging, word sense disambiguation, and missing text prediction. \citet{nehrdich-etal-2024-byt5} extended this approach with ByT5-Sanskrit, a byte-level pretrained model that outperforms tokenizer-based approaches for Sanskrit word segmentation, dependency parsing, and OCR post-correction, demonstrating that byte-level processing is particularly effective for morphologically rich classical languages. Most directly relevant to PaliBench is the MITRA project (\citealp{nehrdich-keutzer-2026-mitra}), which addresses Buddhist canonical languages including Pali in its broader framework, while its 1.74-million parallel sentence-pair corpus is centered on Sanskrit, Chinese, and Tibetan; the project also provides a domain-specific pretrained translation model based on Gemma 2 and a semantic embedding benchmark for classical Buddhist texts. MITRA constitutes the most closely related prior work to PaliBench. PaliBench differs by constructing a multi-reference Pali-to-English benchmark from independent published English translations, foregrounding translation plurality rather than large-scale cross-lingual parallel mining.

\subsection{Multi-Reference Evaluation and Translation Variability}

A central premise of PaliBench is that multiple independent reference translations are not merely desirable but essential for evaluating translation quality in domains where legitimate variability is high. This premise finds growing empirical support. \citet{fomicheva-etal-2020-multi} formalized the insight that translation variability is informative by using multiple MT outputs as reference surrogates, demonstrating a 15\% improvement in metric correlation with human judgments when variability estimates are incorporated. \citet{wu-etal-2024-multiple} provided complementary evidence for literary translation, showing that multiple references with medium to high paraphrastic similarity yield significant improvements in BLEU, COMET, and chrF++ scores, empirically validating the intuition that a single reference penalizes valid alternative renderings. For Pali Buddhist texts, translation variability is not a nuisance parameter but a reflection of interpretive depth. Sujato's modern translations on SuttaCentral illustrate how choices about register, technical terminology, and doctrinal emphasis can produce translations that differ markedly in presentation while still aiming at fidelity to the source \citep{suttacentral-nd}. PaliBench leverages this natural variability by constructing a multi-reference benchmark where the divergence between references serves as both a quality control measure and a window into the evaluation challenges specific to sacred and philosophical texts. By evaluating systems against multiple references simultaneously, PaliBench captures translation adequacy more robustly than single-reference approaches and provides empirical data on how metric behavior shifts in the presence of legitimate interpretive diversity.

\section{Methods}

The method is designed as a reusable benchmark-construction workflow rather than a Pali-only data preparation procedure. Its steps address problems common to classical and religious textual corpora: independently segmented translations, formulaic repetition, partial overlap among translators, copyright-sensitive source materials, and the absence of a single authoritative reference translation. The Pali implementation described below is therefore a worked example of a more general procedure: identify overlapping expert translations, align them to a shared source structure, verify that alignments remain grounded in the original translation files, filter unstable or low-quality units, remove near-duplicates, and evaluate machine outputs against the resulting multi-reference set.

\subsection{Source Corpus and Reference Translations}

PaliBench draws its source material from the Pali Canon (Tipitaka), the foundational scriptural collection of the Theravada Buddhist tradition. The Canon is preserved in Pali, a Middle Indo-Aryan language that serves as the liturgical and scholarly language of Theravada Buddhism. While the Canon comprises three major divisions (the Vinaya Pitaka, Sutta Pitaka, and Abhidhamma Pitaka), the benchmark focuses on selected texts from three collections within the Sutta Pitaka: the Anguttara Nikaya (AN, Numerical Discourses), the Majjhima Nikaya (MN, Middle Length Discourses), and the Samyutta Nikaya (SN, Connected Discourses). These collections were selected because they contain substantial overlap among three established English translators in the source corpus used here. The three reference translators are Bhikkhu Sujato, Bhikkhu Thanissaro, and Bhikkhu Bodhi. Each represents a distinct translation philosophy. Sujato's translations, published under open license on the SuttaCentral platform \citep{suttacentral-nd} and available in JSON format on SuttaCentral's GitHub Repository, often employ modern, accessible English and are segmented with unique identifiers for each segment. Thanissaro's translations, available through dhammatalks.org \citep{thanissaro-nd-suttas}, tend to use more traditional terminology and are formatted as continuous prose. Bodhi's translations, published by Wisdom Publications \citep{bodhi-2000-connected,bodhi-nanamoli-2005-middle,bodhi-2012-numerical}, often adopt a more academic register. The availability of three stylistically diverse translations of the same source texts is what makes the selected corpus well suited to multi-reference machine translation evaluation. The raw corpus comprises 380 source files (100 AN, 39 MN, 241 SN) totaling 22,384 segments, 4,003 passages, and approximately 920,000 tokens. The unit of analysis in PaliBench is the passage---a group of segments sharing the same structural prefix in SuttaCentral's segmentation scheme (e.g., passage mn2:1 contains segments mn2:1.1, mn2:1.2, mn2:1.3). Passages represent coherent semantic units and serve as the primary benchmark unit for both translation and evaluation. Table 1 summarizes the raw corpus statistics.

\begin{table}[t]
\centering
\small
\caption{Raw corpus statistics by collection.}
\label{tab:raw-corpus}
\resizebox{\columnwidth}{!}{%
\begin{tabular}{lrrrr}
\toprule
Collection & Files & Segments & Passages & Tokens (approx.) \\
\midrule
AN & 100 & 4,334 & 872 & 176,000 \\
MN & 39 & 8,395 & 1,348 & 398,000 \\
SN & 241 & 9,655 & 1,783 & 346,000 \\
Total & 380 & 22,384 & 4,003 & 920,000 \\
\bottomrule
\end{tabular}%
}
\end{table}

\subsection{Translation Alignment}

A central methodological challenge in constructing PaliBench is that the three translators segment their texts differently and sometimes make use of abbreviation marks (``\ldots{}'') for repetitive passages---a convention inherited from the Pali manuscript tradition. Sujato's translations are already segmented with unique identifiers on SuttaCentral's GitHub Repository; Thanissaro's and Bodhi's translations, however, exist as continuous prose without segment-level markup. To create a parallel corpus, the translations of Thanissaro and Bodhi must be aligned to Sujato's segment identifiers. We accomplished this alignment using GPT-5-mini \citep{openai-2025-gpt5mini} with structured JSON outputs. For each batch of 15--30 segments, the model received three inputs: the Pali source text (as the semantic anchor determining what content belongs to each segment), Sujato's English translation (as a structural guide indicating segment boundaries), and the full English translation by the target translator (Thanissaro or Bodhi). The model was instructed to extract, for each segment identifier, the corresponding English text from the target translator that expresses the same meaning as the P\=ali. The system prompt enforced several critical rules: output must mirror the structure of the Pali segment (abbreviated Pali yields abbreviated output; fully spelled-out Pali yields fully expanded output); extraction should be minimal and faithful, preferring contiguous substrings from the original; and null values should be returned only when matching content genuinely cannot be found. The complete alignment prompt is reproduced in Appendix B.

Alignment was performed one translator at a time to maximize prompt caching efficiency. The results, shown in Table 2, indicate that 94.1\% of segments were successfully aligned for both translators. The remaining null values are predominantly structural metadata (sutta numbers, colophons, and empty markers) rather than substantive textual content.

\begin{table}[t]
\centering
\small
\caption{Alignment results by collection.}
\label{tab:alignment-results}
\resizebox{\columnwidth}{!}{%
\begin{tabular}{lrrr}
\toprule
Collection & Segments & Thanissaro match & Bodhi match \\
\midrule
AN & 4,334 & 4,036 (93.1\%) & 4,047 (93.4\%) \\
MN & 8,395 & 8,059 (96.0\%) & 8,118 (96.7\%) \\
SN & 9,655 & 8,961 (92.8\%) & 8,910 (92.3\%) \\
Total & 22,384 & 21,056 (94.1\%) & 21,075 (94.1\%) \\
\bottomrule
\end{tabular}%
}
\end{table}

\subsection{Alignment Verification}

Because the alignment pipeline relies on a large language model, automated verification is essential to detect potential hallucinations or extraction errors. For each aligned segment, we checked whether the extracted text could be located in the original source file using a four-stage matching process of increasing tolerance. Verbatim matching requires an exact string match. Normalized matching applies standardization of Unicode characters, quotation marks, dashes, whitespace, and list numbering before comparison. Expanded matching accounts for legitimate reconstruction of abbreviated source text where the translator used ellipsis marks that the alignment process expanded by reference to earlier template passages. Cross-reference matching handles cases where the source text directs the reader to another sutta. Segments that could not be located through any of these methods were flagged as suspicious and excluded from the benchmark. The verification results, presented in Table 3, demonstrate high alignment quality: 95.2\% (Bodhi) and 94.9\% (Thanissaro) of non-null segments match their source verbatim or after minor normalization. Only 0.2\% of segments were flagged as suspicious (approximately 80 segments total), indicating that GPT-5-mini's extraction was highly faithful to the source material.

\begin{table}[t]
\centering
\small
\caption{Alignment verification results (percentage of non-null segments).}
\label{tab:verification-results}
\resizebox{\columnwidth}{!}{%
\begin{tabular}{lrrl}
\toprule
Category & Bodhi & Thanissaro & Description \\
\midrule
Verbatim & 85.1\% & 82.2\% & Exact match in source \\
Normalized & 10.1\% & 12.7\% & Match after text normalization \\
Expanded & 4.5\% & 4.9\% & Legitimate expansion of "\ldots{}" \\
Cross-reference & 0.1\% & 0.0\% & Content from referenced suttas \\
Suspicious & 0.2\% & 0.2\% & Potential error; excluded \\
\bottomrule
\end{tabular}%
}
\end{table}

\subsection{Corpus Filtering and Deduplication}

To ensure benchmark quality, we applied a series of filtering criteria at the passage level. Passage-level filtering was chosen over segment-level filtering because passages are coherent semantic units, and many individual segments are inherently short (e.g., a vocative like ``Bhikkhus'' or a response particle like ``Yes, sir'') where identical translations across translators are legitimate rather than indicative of error. A passage was excluded from the benchmark if any of the following conditions held: 1) incomplete data (any constituent segment has a null translation); 2) verification failure (any segment flagged as suspicious); 3) insufficient length (any translator's rendering is below 100 characters); 4) excessive translation similarity (any pair of translators shares 90\% or greater 3-gram Jaccard similarity, suggesting an alignment copying error); 5) anomalous length ratio (the longest translation exceeds twice the length of the shortest); or 6) internal segment duplication (overlapping or identical content across segments within a passage, indicating an extraction error). These categories are non-exclusive: a single passage may trigger more than one exclusion criterion. Following filtering, we also applied deduplication to address the Pali Canon's extensive formulaic repetition. Identical or near-identical passages frequently appear across multiple texts, and retaining duplicates would inflate the apparent corpus size and provide redundant evaluation data. We compared the Pali text of all passages that passed filtering using 3-gram Jaccard similarity. Where two passages exceeded an 85\% similarity threshold, only one was retained (the first in alphabetical order by passage identifier). Deduplication removed 174 passages (9.3\% of the filtered set). Table 4 summarizes the filtering and deduplication outcomes.

\begin{table}[t]
\centering
\small
\caption{Filtering and deduplication outcomes.}
\label{tab:filtering-deduplication}
\resizebox{\columnwidth}{!}{%
\begin{tabular}{lrr}
\toprule
Stage & Passages affected (non-exclusive) & Passages remaining \\
\midrule
Starting corpus & --- & 4,003 \\
Incomplete data (null segments) & 987 & --- \\
Insufficient length (<100 chars) & 970 & --- \\
Anomalous length ratio (>2.0) & 775 & --- \\
Internal segment duplication & 671 & --- \\
Verification failure & 46 & --- \\
Excessive similarity ($\geq$90\%) & 41 & --- \\
After filtering & --- & 1,874 \\
Near-duplicate Pali passages & 174 & --- \\
Final benchmark dataset & --- & 1,700 \\
\bottomrule
\end{tabular}%
}
\end{table}

Note: Filtering categories are not mutually exclusive; a passage may trigger multiple exclusion criteria.

The final PaliBench dataset comprises 1,700 passages (42.5\% of the original corpus), spanning 8,389 segments and approximately 345,000 tokens. Each passage contains four parallel texts: the Pali source, and English translations by Sujato, Thanissaro, and Bodhi.

\subsection{Evaluation Framework}

Evaluating machine translation of Pali Buddhist texts presents a distinctive challenge: there is no single ``correct'' translation. The three reference translators produce valid but stylistically divergent renderings of the same source, employing different terminological conventions, levels of formality, and interpretive choices. A good machine translation might legitimately differ from all three references while still being accurate. To address this, PaliBench employs a multi-metric evaluation framework that combines four complementary approaches: 1) embedding-based semantic similarity, 2) lexical overlap metrics, 3) a neural translation quality estimator, and 4) surface-level diagnostics. Embedding-based semantic similarity serves as the primary evaluation measure. Each translation (human and machine) is converted into a high-dimensional vector representation using the Qwen3-embedding-8b model \citep{zhang-etal-2025-qwen3}, and cosine similarity is computed between vectors. Because embeddings capture meaning independently of exact wording, this approach tolerates the terminological variation inherent in Pali translation. For instance, ``monks'' and ``mendicants'' receive nearly identical embeddings despite being different words. The key metrics derived from this approach are sim\_best (the highest similarity between the machine translation and any single reference translator, serving as the primary headline score), sim\_centroid (similarity to the mean of all three reference embeddings, measuring proximity to the translators' consensus), and normalized drift (the machine translation's distance from the centroid normalized by the mean inter-translator disagreement, with values above 2.0 flagging outlier translations). Lexical overlap metrics complement the semantic approach by measuring surface-level agreement with the reference translations. We compute BLEU \citep{papineni-etal-2002-bleu} and chrF++ \citep{popovic-2017-chrfpp} using the sacrebleu library \citep{post-2018-call}, with all three human translations provided as references. Corpus-level scores, which aggregate across all 1,700 passages, are reported alongside per-passage means. While embedding similarity captures whether the machine translation conveys the right meaning, lexical metrics reveal whether it does so using the same vocabulary and phrasing as the human translators---a distinction that matters for assessing stylistic alignment. COMET \citep{rei-etal-2022-comet22}, a neural metric trained on human quality judgments from the Workshop on Machine Translation, provides a source-aware evaluation signal. Unlike the preceding metrics, which compare only the machine translation against the human references, COMET also incorporates the Pali source text, enabling it to assess faithfulness to the original. We use the wmt22-comet-da model and compute scores against each reference translator independently, reporting both the mean (comet\_avg) and maximum (comet\_best) across the three references. A known limitation is that COMET was trained on modern language pairs and may extract a weaker signal from Pali source text than from languages in its training data; we therefore treat COMET as a complementary rather than primary metric.

Length ratio and outlier detection provide surface-level diagnostics. The length ratio (machine translation length divided by the mean reference length) flags systematic verbosity or omission, with values near 1.0 indicating well-calibrated output. Outlier detection uses the normalized drift metric described above: passages where normalized drift exceeds 2.0, meaning the machine translation is more than twice as distant from the reference centroid as the average human reference is, are counted as outliers. The outlier rate serves as a measure of translation reliability.

This evaluation design deliberately uses existing expert translations rather than newly collected human judgments. The aim of the present article is to demonstrate a reusable benchmark-construction method and establish reproducible baselines within the space defined by established scholarly translations, not to provide a final philological adjudication of every machine output. Direct expert evaluation remains necessary for validating absolute translation quality, but assembling a sufficiently large and consistent panel is difficult in this domain: Pali expertise is scarce, and the subset of scholars able to assess Pali source passages while also making fine-grained judgments about English translation quality is smaller still. For this reason, the present study treats the three published reference translations as the available expert signal and evaluates model outputs by their relation to that multi-reference space.

\subsection{Model Selection and Translation Protocol}

We evaluated ten large language models on the full PaliBench test set, encompassing both proprietary and open-weight systems. The models were GPT-5.2 (OpenAI), Gemini 3 Pro and Gemini 3 Flash (Google), Claude Opus 4.5 (Anthropic), Qwen3 235B (Alibaba), DeepSeek v3.2 (DeepSeek), Grok 4.1 Fast (xAI), GLM-4.7 (Zhipu AI), Kimi K2.5 (Moonshot AI), and LLaMA 3.3 70B (Meta). All models were accessed via the OpenRouter API \citep{openrouter-nd-api} during January--February 2026. The exact OpenRouter identifiers were openai/gpt-5.2, google/gemini-3-pro-preview, google/gemini-3-flash-preview, anthropic/claude-opus-4.5, qwen/qwen3-235b-a22b-2507, deepseek/deepseek-v3.2, x-ai/grok-4.1-fast, z-ai/glm-4.7, moonshotai/kimi-k2.5, and meta-llama/llama-3.3-70b-instruct. Each passage was evaluated independently, without surrounding sutta context. For API efficiency, multiple passages were sometimes batched in the same JSON request, but batched passages were independent evaluation items rather than surrounding discourse context; no examples or discourse metadata were provided, and outputs were scored at the passage level. No few-shot examples of Pali translation were provided; the task relies entirely on the model's pre-existing knowledge of Pali. This zero-shot, passage-level design follows standard machine translation evaluation practice (as in WMT and FLORES benchmarks), enabling fair comparison across models by ensuring all receive identical input and isolating differences in Pali linguistic competence rather than context utilization strategies. A limitation of this design is that human translators work with full sutta context, and certain passages may be ambiguous without surrounding material. However, PaliBench's filtering ensures that all benchmark passages are semantic units of at least 100 characters, and the highly formulaic nature of Pali canonical texts (with standard openings, stock phrases, and clear structural conventions) means that most passages are interpretable in isolation.

\section{Results}

\subsection{Inter-Translator Baseline}

Before presenting the machine translation results, it is necessary to establish how the three human translators relate to each other, as this defines the interpretive frame for all subsequent evaluation. Table 5 presents the pairwise semantic similarity between the reference translators, computed as the corpus-wide mean cosine similarity across all 1,700 passages.

\begin{table}[t]
\centering
\small
\caption{Pairwise semantic similarity between reference translators (corpus-wide mean cosine similarity).}
\label{tab:translator-similarity}
\resizebox{\columnwidth}{!}{%
\begin{tabular}{lrrrr}
\toprule
Translator pair & Mean & Std & Min & Max \\
\midrule
Thanissaro $\leftrightarrow$ Bodhi & 0.891 & 0.061 & 0.503 & 0.994 \\
Sujato $\leftrightarrow$ Bodhi & 0.876 & 0.056 & 0.552 & 0.980 \\
Sujato $\leftrightarrow$ Thanissaro & 0.859 & 0.064 & 0.509 & 0.979 \\
\bottomrule
\end{tabular}%
}
\end{table}

The pairwise similarities reveal a clear geometric structure among the three translators. Bodhi occupies the position closest to the semantic centroid of the group, with a mean similarity to the centroid of 0.963 and the lowest drift (0.037). Thanissaro is intermediate (centroid similarity 0.957, drift 0.043), while Sujato is the most semantically distinctive translator (centroid similarity 0.952, drift 0.048), with the lowest pairwise similarity to both other translators. The mean human drift (the average distance of the three translators from their own centroid) is 0.043 ($\pm$ 0.019) across all passages. This value serves as the baseline against which machine translation drift is normalized: a machine translation with normalized drift below 1.0 falls within the typical range of human disagreement, while values above 2.0 indicate output that diverges from all three references more than the translators diverge from each other.

\subsection{Overall Model Performance}

Table 6 reports the primary benchmark results for all ten evaluated models, ranked by their best semantic similarity to any of the three reference translations.

\begin{table*}[t]
\centering
\small
\caption{Primary benchmark results. Models ranked by sim\_best.}
\label{tab:primary-results}
\resizebox{\textwidth}{!}{%
\begin{tabular}{rlrrrrrr}
\toprule
Rank & Model & sim\_best & chrF++ & BLEU & COMET\_avg & Length & Outliers \\
\midrule
1 & Gemini 3 Pro & 0.946 & 68.5 & 63.3 & 0.729 & 1.006 & 3.4\% \\
2 & Gemini 3 Flash & 0.944 & 65.9 & 57.9 & 0.731 & 1.060 & 3.9\% \\
3 & Claude Opus 4.5 & 0.940 & 65.6 & 56.2 & 0.724 & 1.027 & 5.5\% \\
4 & DeepSeek v3.2 & 0.939 & 64.1 & 51.3 & 0.718 & 1.068 & 6.7\% \\
5 & Kimi K2.5 & 0.934 & 61.2 & 47.5 & 0.707 & 1.068 & 7.9\% \\
6 & GPT-5.2 & 0.933 & 59.1 & 42.5 & 0.709 & 1.117 & 7.5\% \\
7 & GLM-4.7 & 0.920 & 57.1 & 43.7 & 0.703 & 1.050 & 15.0\% \\
8 & Qwen3 235B & 0.918 & 55.7 & 40.6 & 0.706 & 1.052 & 19.1\% \\
9 & Grok 4.1 Fast & 0.915 & 54.8 & 40.1 & 0.696 & 1.043 & 18.1\% \\
10 & LLaMA 3.3 70B & 0.888 & 48.1 & 34.4 & 0.681 & 1.007 & 40.3\% \\
\bottomrule
\end{tabular}%
}
\end{table*}

Note: COMET\_avg is the mean COMET score across the three reference translators.

The results reveal a tiered structure. The top tier (Gemini 3 Pro, Gemini 3 Flash, Claude Opus 4.5, and DeepSeek v3.2) achieves sim\_best scores above 0.939 and outlier rates below 7\%, indicating translations that are semantically comparable to human references for the vast majority of passages. A middle tier (Kimi K2.5, GPT-5.2, GLM-4.7, Qwen3 235B, and Grok 4.1 Fast) scores between 0.915 and 0.934, with outlier rates of 7--19\%. LLaMA 3.3 70B forms a bottom tier with a sim\_best of 0.888 and a 40.3\% outlier rate. Among the top-tier models, Gemini 3 Pro combines the highest semantic similarity (0.946) with the closest length match to human references (1.006), the highest lexical overlap scores (chrF++ 68.5, BLEU 63.3), and the lowest outlier rate (3.4\%). GPT-5.2 presents an informative contrast: it ranks sixth on sim\_best (0.933) but produces the longest translations of any model (length ratio 1.117), suggesting that it captures the meaning of the Pali source effectively but does so with more verbose, paraphrastic renderings that diverge from the phrasing of the human references.

\subsection{Translator Alignment Patterns}

For each passage, we identified which of the three reference translators the machine translation most closely resembles (by cosine similarity). The resulting distribution, shown in Table 7, reveals two distinct alignment strategies across the ten models.

\begin{table}[t]
\centering
\small
\caption{Closest translator distribution (percentage of 1,700 passages where each translator is the nearest match).}
\label{tab:closest-translator}
\resizebox{\columnwidth}{!}{%
\begin{tabular}{lrrr}
\toprule
Model & Bodhi & Sujato & Thanissaro \\
\midrule
DeepSeek v3.2 & 73.5\% & 7.8\% & 18.8\% \\
Kimi K2.5 & 72.4\% & 9.3\% & 18.4\% \\
Qwen3 235B & 71.4\% & 10.1\% & 18.5\% \\
GPT-5.2 & 69.8\% & 9.4\% & 20.8\% \\
Gemini 3 Flash & 69.1\% & 17.3\% & 13.6\% \\
Grok 4.1 Fast & 64.0\% & 12.6\% & 23.4\% \\
GLM-4.7 & 63.4\% & 13.6\% & 23.0\% \\
LLaMA 3.3 70B & 62.6\% & 13.5\% & 23.8\% \\
Gemini 3 Pro & 44.5\% & 42.1\% & 13.4\% \\
Claude Opus 4.5 & 39.4\% & 48.6\% & 11.9\% \\
\bottomrule
\end{tabular}%
}
\end{table}

Eight of the ten models are Bodhi-dominant, with 63--74\% of passages most closely matching Bodhi's translations. This is consistent with the inter-translator baseline showing Bodhi as the semantic centroid: any model producing a consensus-style translation will naturally gravitate toward the centroid translator. Two models deviate from this pattern. Claude Opus 4.5 is the only model for which Sujato is the most frequently matched translator (48.6\%), suggesting that its training data may have included substantial exposure to Sujato's modern, accessible renderings or a general preference for modern accessible phrasing. Gemini 3 Pro exhibits a near-even split between Sujato (42.1\%) and Bodhi (44.5\%), indicating a balanced synthesis of both translators' approaches. Notably, both models that show elevated Sujato alignment rank in the top three overall, suggesting that proximity to the more distinctive Sujato style does not penalize performance under the multi-reference evaluation framework.

\subsection{Translation Consistency and Reliability}

The outlier rate (the percentage of passages where the machine translation's normalized drift exceeds 2.0) serves as a measure of translation reliability, capturing how often a model produces output that falls substantially outside the envelope of human translator variation. For the top-tier models, outlier rates range from 3.4\% (Gemini 3 Pro) to 6.7\% (DeepSeek v3.2), meaning that over 93\% of their translations fall within the bounds defined by inter-translator disagreement. In the middle tier, outlier rates rise to 7--19\%, and LLaMA 3.3 70B's 40.3\% outlier rate indicates that a substantial fraction of its output diverges from all three references in ways that would warrant manual review.

\subsection{Cross-Metric Concordance}

To assess the robustness of the model rankings, Table 8 presents each model's rank across six evaluation dimensions: semantic similarity (sim\_best), chrF++, BLEU, COMET\_avg, length calibration (absolute deviation from 1.0), and outlier rate. The mean rank is the unweighted average across these dimensions.

\begin{table*}[t]
\centering
\small
\caption{Cross-metric rankings (1 = best). Mean rank is the unweighted average across valid dimensions.}
\label{tab:cross-metric-rankings}
\resizebox{\textwidth}{!}{%
\begin{tabular}{lrrrrrrr}
\toprule
Model & sim\_best & chrF++ & BLEU & COMET\_avg & Length & Outliers & Mean \\
\midrule
Gemini 3 Pro & 1 & 1 & 1 & 2 & 1 & 1 & 1.2 \\
Gemini 3 Flash & 2 & 2 & 2 & 1 & 7 & 2 & 2.7 \\
Claude Opus 4.5 & 3 & 3 & 3 & 3 & 3 & 3 & 3.0 \\
DeepSeek v3.2 & 4 & 4 & 4 & 4 & 8 & 4 & 4.7 \\
Kimi K2.5 & 5 & 5 & 5 & 6 & 9 & 6 & 6.0 \\
GPT-5.2 & 6 & 6 & 7 & 5 & 10 & 5 & 6.5 \\
GLM-4.7 & 7 & 7 & 6 & 8 & 5 & 7 & 6.7 \\
Qwen3 235B & 8 & 8 & 8 & 7 & 6 & 9 & 7.7 \\
Grok 4.1 Fast & 9 & 9 & 9 & 9 & 4 & 8 & 8.0 \\
LLaMA 3.3 70B & 10 & 10 & 10 & 10 & 2 & 10 & 8.7 \\
\bottomrule
\end{tabular}%
}
\end{table*}

The rankings exhibit strong concordance across metric categories. The sim\_best, chrF++, and BLEU rankings are nearly identical for the top six models, and COMET\_avg rankings largely mirror the pattern, with only minor reorderings (Gemini 3 Flash edges ahead of Gemini 3 Pro on COMET\_avg while trailing on all other metrics). Length calibration introduces the most variation, as it measures a qualitatively different property (verbosity rather than accuracy), but even here the top-ranked model on sim\_best also achieves the best length calibration. The strong agreement across semantic, lexical, and neural metrics strengthens confidence that the observed rankings reflect genuine differences in Pali translation competence rather than artifacts of any single evaluation methodology. One notable divergence is GPT-5.2, which ranks fifth on COMET\_avg but sixth on sim\_best and tenth on length calibration, reinforcing the observation from Section 4.2 that it produces semantically adequate but stylistically verbose translations. This pattern illustrates the value of the multi-metric approach: relying on any single metric would obscure meaningful differences in translation behavior.

\section{Discussion}

\subsection{The Geometry of Translation and the Risk of Interpretive Flattening}

Perhaps the most consequential finding of this study is not which model scored highest, but what the translator alignment patterns reveal about how large language models relate to existing traditions of Pali scholarship. The inter-translator baseline established that Bodhi occupies the semantic centroid of the three reference translators, while Sujato is the most distinctive voice. When eight of ten models gravitate toward Bodhi's register---with 63--74\% of passages most closely matching his renderings---this is likely not merely a statistical artifact of centroid attraction. It may signal that current LLMs, when translating Pali, tend to converge on a consensus-style output that resembles the most conservative, academically conventional translation available. This convergence has implications that extend beyond translation quality metrics. In Pali Buddhist studies, the differences between translators are not noise to be averaged away; they are substantive interpretive choices that reflect distinct understandings of the source tradition. Sujato's rendering of bhikkhu as ``mendicant'' rather than Bodhi's ``bhikkhu'' or Thanissaro's ``monk'' is not a stylistic preference. Instead, it is an argument about how to make the Pali term legible to a modern audience without, for example, importing associations from other religious traditions. Similarly, choices about how to render key doctrinal terms like dukkha (suffering, stress, unsatisfactoriness), sa\.nkh\=ara (formations, fabrications, volitional activities), or vi\~n\~n\=a\d{n}a (consciousness, cognizance, discernment) carry interpretive weight that shapes how readers understand fundamental Buddhist concepts. If LLM translation at scale defaults to centroid-style output, the risk is an inadvertent flattening of the interpretive diversity that characterizes living scholarly traditions. A reader encountering only AI-generated translations of untranslated texts would receive a single homogenized voice where multiple legitimate perspectives exist. The two exceptions in our data, Claude Opus 4.5 and Gemini 3 Pro, both of which show elevated alignment with Sujato's more distinctive register, suggest that this convergence is not inevitable: models whose training data includes greater representation of diverse translation traditions may produce output that preserves interpretive plurality. This finding has practical implications for how training corpora are curated for classical language tasks, and for how AI-assisted translation tools might be designed to surface rather than suppress the multiplicity of valid renderings.

\subsection{What Top-Tier Performance Means, and What It Does Not}

The headline result (that the best-performing models achieve sim\_best scores above 0.94, with very few outliers) invites optimism about the state of LLM translation for classical languages. These numbers indicate that, for the majority of benchmark passages, the semantic content of the machine translation is as close to at least one human reference as the human references are to each other. This is a meaningful threshold: it suggests that top-tier models may have internalised substantial knowledge of Pali grammar, vocabulary, and canonical idiom. However, several caveats temper this interpretation. First, the benchmark evaluates passage-level translation without document context. Human translators work with full suttas and make choices that reflect the arc of an entire discourse; the benchmark cannot assess this dimension of translation competence. Second, the evaluation is purely reference-based: it measures how closely the machine output resembles existing human translations, not whether the machine output is independently correct. A machine translation that reproduces a debatable interpretive choice from a reference translator would score well on similarity metrics without being accurate in any deeper philological sense. Third, the outlier rates across models indicate that even the best systems produce a non-trivial proportion of passages where the output diverges substantially from all three references. These passages would require expert review before being considered reliable. The COMET results warrant particular caution. While COMET is the state-of-the-art neural evaluation metric in machine translation and has been the primary metric at the Workshop on Machine Translation since 2022, it was trained on modern language pairs (English, German, Chinese, and others), not Pali. The source encoder's signal for Pali is likely weaker than for languages in its training data, meaning that COMET's source-awareness (its key advantage over reference-only metrics) may be attenuated in this setting. We therefore treated COMET as a complementary rather than primary metric. The strong concordance between COMET rankings and the other metric families suggests that its reference-side and hypothesis-side signals remain informative, but the absolute COMET scores should be interpreted with caution relative to benchmarks on modern language pairs.

\subsection{A Blueprint for Classical Language Translation Benchmarks}

The Pali case should be understood as a worked example of a broader digital humanities method. PaliBench was designed for Pali-to-English translation, but its workflow is portable to any classical or low-resource language for which multiple independent expert translations of the same source corpus exist. The core design (LLM-assisted alignment of independently segmented translations, multi-stage verification to detect extraction errors, passage-level filtering for benchmark quality, and multi-metric evaluation against multiple references) addresses challenges common to independently produced translation sets. The number of reference translators need not be three; the methodology generalizes to any number of references, with more references providing a richer semantic centroid and more robust evaluation. Classical languages are particularly well suited to this approach because they present a shared set of characteristics that make single-reference evaluation inadequate: archaic grammar and vocabulary for which no modern consensus translation exists, extensive formulaic repetition that complicates automated processing, technical terminology with contested renderings across scholarly traditions, and the fundamental reality that translation of ancient texts is always an act of interpretation rather than transcription. The multi-metric framework directly addresses these characteristics by combining semantic embeddings that tolerate stylistic variation with lexical metrics that reward phrasing precision and neural metrics that assess source faithfulness. Several classical languages present immediate opportunities for benchmark construction using PaliBench's pipeline. Sanskrit texts (including the Bhagavad Gita and the Upanishads) have multiple acclaimed English translations \citep[e.g.,][]{olivelle-1996-upanisads,roebuck-2003-upanishads,van-buitenen-1981-bhagavadgita}, and the extensive overlap between the Pali and Sanskrit Buddhist canons makes this a particularly natural extension. The Confucian and Daoist classics of Classical Chinese (the Analects, the Dao De Jing, the Zhuangzi) and Chinese Buddhist texts have numerous competing English translations. Latin philosophical, theological, and literary texts (Seneca, Augustine, Virgil) each have multiple modern English translations spanning centuries of scholarly tradition. Ancient Greek literature and philosophy, Classical Arabic philosophical and religious texts, Biblical Hebrew, and Classical Tibetan all offer comparable multi-reference resources. The key prerequisite in each case is the existence of multiple independent translations of sufficient scope to construct a statistically meaningful test set.

\subsection{Implications for Access to Classical Texts}

The Pali Canon contains 10,000-plus suttas \citep{access-to-insight-nd}, and only a fraction are available in multiple independent English translations. Many canonical texts and many modern target-language pairings remain without comparable expert translation coverage. The situation is analogous for other classical language corpora: the volume of extant texts vastly exceeds the capacity of the small number of scholars qualified to translate them. Large language models offer the potential to expand access to these texts by producing serviceable translations at scale---not only into English, but into any modern language. A Vietnamese Buddhist practitioner seeking to read a Pali sutta not yet available in Vietnamese, or a Spanish-speaking student of Stoicism wanting to engage with a Latin text unavailable in Spanish, could benefit from LLM-generated translations that, while imperfect, provide meaningful access where none previously existed. This is not a replacement scenario: no human translation exists to be replaced. It is an expansion of access that would otherwise require years or decades of scholarly labor. The benchmark results, however, underscore that quality assurance is critical. Even the best-performing model produces outlier translations for 3.4\% of passages, and all models exhibit non-trivial variance in quality. For religious, philosophical, and legal texts, where mistranslation can propagate doctrinal misunderstanding or misrepresent intellectual traditions, unchecked AI translation poses real risks. Benchmarks like PaliBench contribute to the quality assurance infrastructure needed to deploy LLM translation responsibly: they provide standardized evaluation, identify systematic failure modes (outliers, excessive paraphrasing, terminology drift), and establish performance baselines against which model improvements can be measured. For professional translators, the results suggest a complementary role for LLM-generated output. The benchmark data indicate that top-tier models capture the correct semantic content for most of the passages, meaning that a human expert reviewing an AI draft would spend most of their effort on stylistic refinement and edge-case correction rather than wholesale retranslation. This represents a meaningful productivity improvement for a field where a single text can require weeks of careful scholarly work. The 3--40\% outlier rates across models reinforce that human review remains essential; AI translations at current quality levels are best understood as informed first drafts that require expert oversight for doctrinal accuracy, contextual sensitivity, and the literary qualities that distinguish a scholarly translation from a functional one. The goal is augmentation, not replacement, and the benchmark itself is built on the foundation of human expertise, as the decades of scholarly work by Bhikkhu Sujato, Bhikkhu Thanissaro, and Bhikkhu Bodhi provide the reference translations against which all AI output is measured.

\subsection{Limitations}

Several limitations should be noted. First, PaliBench covers only three of the five Nikayas in the Sutta Pitaka (AN, MN, SN), excluding the Digha Nikaya and Khuddaka Nikaya as well as the Vinaya and Abhidhamma Pitakas. Extension to these collections is constrained by the availability of parallel translations from all three reference translators. Second, the selection of three translators, while enabling multi-reference evaluation, does not exhaust the available translations of the Pali Canon; other scholars \citep[e.g.,][]{walshe-1995-long,nanamoli-bodhi-1995-middle} have produced respected translations, however, translations overlapping with the three reference translators were not available in large enough numbers to include them. Third, the filtering pipeline retains only 42.5\% of original passages, and the deduplication step removes repetitive but authentic canonical content. While these steps are necessary for benchmark quality, they mean that the test set is not representative of the full Pali Canon, particularly its extensive use of formulaic repetition. Fourth, the passage-level translation design, while standard in MT evaluation, diverges from how human translators work and cannot assess document-level coherence. Fifth, due to practical constraints, the benchmark lacks human evaluation: all metrics are automated, and the correlation between these automated scores and expert judgments of Pali translation quality has not been empirically validated. Finally, the alignment pipeline relies on a large language model (GPT-5-mini), and while our verification pipeline was designed to mitigate this risk, subtle alignment errors below the detection threshold of our verification methodology cannot be entirely excluded.

\section{Conclusion}

This paper has presented PaliBench as both a multi-reference benchmark for evaluating machine translation of Pali Buddhist canonical texts and a reusable method for constructing translation benchmarks in other classical language settings. The benchmark comprises 1,700 parallel passages drawn from three collections of the Sutta Pitaka, each with four aligned texts: the Pali source and English translations by three established scholar-translators. A five-stage pipeline (alignment, verification, filtering, deduplication, and multi-metric evaluation) transforms these raw materials into an evaluation framework that accommodates the inherent plurality of valid Pali translations.

The empirical evaluation of ten large language models demonstrates how such a benchmark can be used: the best current systems achieve semantic similarity scores above 0.94 relative to human references, with over 93\% of passages falling within the bounds of normal inter-translator variation, while all models still produce outlier translations at non-trivial rates (3--40\%). Translator alignment analysis further shows that most models converge on centroid-style output that may flatten the interpretive diversity characteristic of Pali scholarship. These findings have practical implications for the deployment of LLM translation in classical language contexts, but the central contribution is methodological. PaliBench shows that existing scholarly translations can be converted into benchmark infrastructure without treating any single translation as definitive. The same pipeline can be adapted to Sanskrit, Classical Chinese, Latin, Ancient Greek, Classical Arabic, Biblical Hebrew, Classical Tibetan, and other traditions where enough independent reference translations exist. As large language models continue to improve in their handling of low-resource languages, standardized benchmarks grounded in scholarly translation traditions will be increasingly important for ensuring that AI-generated translations meet the standards of accuracy and interpretive care that these texts demand.

\section*{Data and Code Availability}

A public reproducibility package is made available at https://github.com/MateMetzger/palibench. The package includes the pipeline and evaluation code, translation and alignment prompts, corpus-construction metadata, passage and segment identifiers, text-length metadata, aggregate model results, per-passage metric scores, and model-generated translation outputs. It intentionally excludes the full benchmark file containing Pali source passages and aligned human reference translations, raw source files, aligned reference files, reference embeddings, and outlier-review files that quote full passages. The underlying source and reference translations are freely accessible online from the cited providers, but they are distributed under different terms: Sujato's SuttaCentral translations are released under Creative Commons Zero (CC0), Thanissaro's translations are available under Creative Commons Attribution-NonCommercial 4.0, and Bodhi's Wisdom Publications translations are available under Creative Commons Attribution-NonCommercial-NoDerivs 3.0 Unported. Because the latter license includes a NoDerivs condition, and because making aligned full-text extraction publicly available may constitute derivative-work, the public package excludes all human translation text rather than treating the three translators differently. However, all three translations are freely available online. Researchers may reconstruct the alignment locally for personal scholarly use as permitted by the applicable licenses, using the provided passage manifest, prompts, pipeline scripts, and metadata.

\section*{Competing Interests}

The authors have no competing interests to declare.

\section*{Author Contributions}

M\'at\'e Metzger: Conceptualization, Data curation, Formal analysis, Investigation, Methodology, Software, Validation, Writing - original draft, Writing - review and editing, Project administration.

Nadnapang Phophichit: Conceptualization, Supervision, Validation, Writing - review and editing.

\section*{Generative AI Use}

Generative AI was used in two distinct ways in this study. First, large language models were used as part of the research object and method: GPT-5-mini was used for alignment of independently segmented translations as described in Section 3.2, and the evaluated systems generated the machine translations analyzed in the benchmark. Second, generative AI tools were used during manuscript and software preparation to assist with code generation, language polishing, structural revision, and consistency checking. Generative AI was not used to autonomously generate the manuscript, independently analyze the results, fabricate data or citations, or determine the study's conclusions. All AI-assisted outputs were reviewed, verified, and revised by the authors, who take full responsibility for the content of the article, the reported results, and the accompanying code.

\nocite{*}
\bibliography{custom}

@misc{access-to-insight-nd,
  author = {{Access to Insight}},
  title = {Frequently Asked Questions About Access to Insight},
  year = {n.d.},
  note = {Accessed April 30, 2026},
  url = {https://www.accesstoinsight.org/faq.html}
}

@article{assael-etal-2022-restoring,
  author = {Assael, Yannis and Sommerschield, Thea and Shillingford, Brendan and others},
  title = {Restoring and Attributing Ancient Texts Using Deep Neural Networks},
  journal = {Nature},
  volume = {603},
  number = {7900},
  pages = {280--283},
  year = {2022},
  doi = {10.1038/s41586-022-04448-z}
}

@misc{bamman-burns-2020-latin,
  author = {Bamman, David and Burns, Patrick J.},
  title = {Latin {BERT}: A Contextual Language Model for Classical Philology},
  year = {2020},
  note = {arXiv preprint},
  doi = {10.48550/arXiv.2009.10053}
}

@book{bodhi-2000-connected,
  author = {Bodhi, Bhikkhu},
  title = {The Connected Discourses of the Buddha: A Translation of the Sa{\d m}yutta Nik{\=a}ya},
  publisher = {Wisdom Publications},
  year = {2000}
}

@book{bodhi-nanamoli-2005-middle,
  author = {Bodhi, Bhikkhu and {\~N}{\=a}{\d n}amoli, Bhikkhu},
  title = {The Middle Length Discourses of the Buddha: A Translation of the Majjhima Nik{\=a}ya},
  edition = {3rd},
  publisher = {Wisdom Publications in association with the Barre Center for Buddhist Studies},
  year = {2005}
}

@book{bodhi-2012-numerical,
  author = {Bodhi, Bhikkhu},
  title = {The Numerical Discourses of the Buddha: A Translation of the A{\.n}guttara Nik{\=a}ya},
  publisher = {Wisdom Publications},
  year = {2012}
}

@inproceedings{fomicheva-etal-2020-multi,
  author = {Fomicheva, Marina and Specia, Lucia and Guzm{\'a}n, Francisco},
  title = {Multi-Hypothesis Machine Translation Evaluation},
  booktitle = {Proceedings of the 58th Annual Meeting of the Association for Computational Linguistics},
  pages = {1218--1232},
  year = {2020},
  doi = {10.18653/v1/2020.acl-main.113}
}

@inproceedings{freitag-etal-2022-results,
  author = {Freitag, Markus and Rei, Ricardo and Mathur, Nitika and others},
  title = {Results of {WMT22} Metrics Shared Task: Stop Using {BLEU} -- Neural Metrics Are Better and More Robust},
  booktitle = {Proceedings of the Seventh Conference on Machine Translation},
  pages = {46--68},
  year = {2022},
  doi = {10.18653/v1/2022.wmt-1.2}
}

@article{goyal-etal-2022-flores,
  author = {Goyal, Naman and Gao, Cynthia and Chaudhary, Vishrav and others},
  title = {The {FLORES}-101 Evaluation Benchmark for Low-Resource and Multilingual Machine Translation},
  journal = {Transactions of the Association for Computational Linguistics},
  volume = {10},
  pages = {522--538},
  year = {2022},
  doi = {10.1162/tacl_a_00474}
}

@article{guerreiro-etal-2024-xcomet,
  author = {Guerreiro, Nuno M. and Rei, Ricardo and van Stigt, Daan and Coheur, Luisa and Colombo, Pierre and Martins, Andr{\'e} F. T.},
  title = {{xCOMET}: Transparent Machine Translation Evaluation through Fine-Grained Error Detection},
  journal = {Transactions of the Association for Computational Linguistics},
  volume = {12},
  pages = {979--995},
  year = {2024},
  doi = {10.1162/tacl_a_00683}
}

@incollection{hellwig-2015-morphological,
  author = {Hellwig, Oliver},
  title = {Morphological Disambiguation of Classical Sanskrit},
  booktitle = {Systems and Frameworks for Computational Morphology},
  editor = {Mahlow, Cerstin and Piotrowski, Michael},
  pages = {41--59},
  publisher = {Springer},
  year = {2015},
  doi = {10.1007/978-3-319-23980-4_3}
}

@misc{hendy-etal-2023-gpt,
  author = {Hendy, Amr and Abdelrehim, Mohamed and Sharaf, Amr and others},
  title = {How Good Are {GPT} Models at Machine Translation? A Comprehensive Evaluation},
  year = {2023},
  note = {arXiv preprint},
  doi = {10.48550/arXiv.2302.09210}
}

@misc{jiao-etal-2023-chatgpt,
  author = {Jiao, Wenxiang and Wang, Wenxuan and Huang, Jen-tse and Wang, Xing and Tu, Zhaopeng},
  title = {Is {ChatGPT} a Good Translator? Yes with {GPT}-4 as the Engine},
  year = {2023},
  note = {arXiv preprint},
  doi = {10.48550/arXiv.2301.08745}
}

@inproceedings{kocmi-federmann-2023-gemba,
  author = {Kocmi, Tom and Federmann, Christian},
  title = {{GEMBA-MQM}: Detecting Translation Quality Error Spans with {GPT}-4},
  booktitle = {Proceedings of the Eighth Conference on Machine Translation},
  pages = {768--775},
  year = {2023},
  doi = {10.18653/v1/2023.wmt-1.64}
}

@article{lavie-denkowski-2009-meteor,
  author = {Lavie, Alon and Denkowski, Michael J.},
  title = {The Meteor Metric for Automatic Evaluation of Machine Translation},
  journal = {Machine Translation},
  volume = {23},
  number = {2--3},
  pages = {105--115},
  year = {2009},
  doi = {10.1007/s10590-009-9059-4}
}

@inproceedings{nehrdich-etal-2024-byt5,
  author = {Nehrdich, Sebastian and Hellwig, Oliver and Keutzer, Kurt},
  title = {One Model Is All You Need: {ByT5-Sanskrit}, a Unified Model for Sanskrit {NLP} Tasks},
  booktitle = {Findings of the Association for Computational Linguistics: EMNLP 2024},
  pages = {13742--13751},
  year = {2024},
  doi = {10.18653/v1/2024.findings-emnlp.805}
}

@misc{nehrdich-keutzer-2026-mitra,
  author = {Nehrdich, Sebastian and Keutzer, Kurt},
  title = {{MITRA}: A Large-Scale Parallel Corpus and Multilingual Pretrained Language Model for Machine Translation and Semantic Retrieval for P{\=a}li, Sanskrit, Buddhist Chinese, and Tibetan},
  year = {2026},
  note = {arXiv preprint, arXiv:2601.06400},
  doi = {10.48550/arXiv.2601.06400}
}

@misc{nllb-etal-2022-no,
  author = {{NLLB Team} and Costa-juss{\`a}, Marta R. and Cross, James and others},
  title = {No Language Left Behind: Scaling Human-Centered Machine Translation},
  year = {2022},
  note = {arXiv preprint, arXiv:2207.04672}
}

@article{nllb-team-2024-scaling,
  author = {{NLLB Team}},
  title = {Scaling Neural Machine Translation to 200 Languages},
  journal = {Nature},
  volume = {630},
  number = {8018},
  pages = {841--846},
  year = {2024},
  doi = {10.1038/s41586-024-07335-x}
}

@book{nanamoli-bodhi-1995-middle,
  author = {{\~N}{\=a}{\d n}amoli, Bhikkhu and Bodhi, Bhikkhu},
  title = {The Middle Length Discourses of the Buddha: A Translation of the Majjhima Nik{\=a}ya},
  publisher = {Wisdom Publications},
  year = {1995}
}

@book{olivelle-1996-upanisads,
  author = {Olivelle, Patrick},
  title = {Upani{\d s}ads},
  publisher = {Oxford University Press},
  year = {1996}
}

@misc{openai-2025-gpt5mini,
  author = {{OpenAI}},
  title = {{GPT}-5 mini},
  year = {2025},
  note = {OpenAI API documentation},
  url = {https://platform.openai.com/docs/models/gpt-5-mini}
}

@misc{openrouter-nd-api,
  author = {{OpenRouter}},
  title = {{API} Reference},
  year = {n.d.},
  note = {Accessed April 30, 2026},
  url = {https://openrouter.ai/docs/api/reference}
}

@inproceedings{papineni-etal-2002-bleu,
  author = {Papineni, Kishore and Roukos, Salim and Ward, Todd and Zhu, Wei-Jing},
  title = {{BLEU}: A Method for Automatic Evaluation of Machine Translation},
  booktitle = {Proceedings of the 40th Annual Meeting of the Association for Computational Linguistics},
  pages = {311--318},
  year = {2002},
  doi = {10.3115/1073083.1073135}
}

@inproceedings{popovic-2015-chrf,
  author = {Popovi{\'c}, Maja},
  title = {{chrF}: Character N-Gram F-Score for Automatic {MT} Evaluation},
  booktitle = {Proceedings of the Tenth Workshop on Statistical Machine Translation},
  pages = {392--395},
  year = {2015},
  doi = {10.18653/v1/W15-3049}
}

@inproceedings{popovic-2017-chrfpp,
  author = {Popovi{\'c}, Maja},
  title = {{chrF++}: Words Helping Character N-Grams},
  booktitle = {Proceedings of the Second Conference on Machine Translation},
  pages = {612--618},
  year = {2017},
  doi = {10.18653/v1/W17-4770}
}

@inproceedings{post-2018-call,
  author = {Post, Matt},
  title = {A Call for Clarity in Reporting {BLEU} Scores},
  booktitle = {Proceedings of the Third Conference on Machine Translation},
  pages = {186--191},
  year = {2018},
  doi = {10.18653/v1/W18-6319}
}

@inproceedings{rei-etal-2020-comet,
  author = {Rei, Ricardo and Stewart, Craig and Farinha, Ana C. and Lavie, Alon},
  title = {{COMET}: A Neural Framework for {MT} Evaluation},
  booktitle = {Proceedings of the 2020 Conference on Empirical Methods in Natural Language Processing},
  pages = {2685--2702},
  year = {2020},
  doi = {10.18653/v1/2020.emnlp-main.213}
}

@inproceedings{rei-etal-2022-comet22,
  author = {Rei, Ricardo and de Souza, Jos{\'e} G. C. and Alves, Duarte and others},
  title = {{COMET}-22: {Unbabel-IST} 2022 Submission for the Metrics Shared Task},
  booktitle = {Proceedings of the Seventh Conference on Machine Translation (WMT)},
  pages = {578--585},
  year = {2022},
  doi = {10.18653/v1/2022.wmt-1.52}
}

@book{roebuck-2003-upanishads,
  author = {Roebuck, Valerie J.},
  title = {The Upanishads},
  publisher = {Penguin Books},
  year = {2003}
}

@inproceedings{sellam-etal-2020-bleurt,
  author = {Sellam, Thibault and Das, Dipanjan and Parikh, Ankur},
  title = {{BLEURT}: Learning Robust Metrics for Text Generation},
  booktitle = {Proceedings of the 58th Annual Meeting of the Association for Computational Linguistics},
  pages = {7881--7892},
  year = {2020},
  doi = {10.18653/v1/2020.acl-main.704}
}

@article{sommerschield-etal-2023-survey,
  author = {Sommerschield, Thea and Assael, Yannis and Pavlopoulos, John and others},
  title = {Machine Learning for Ancient Languages: A Survey},
  journal = {Computational Linguistics},
  volume = {49},
  number = {3},
  pages = {703--747},
  year = {2023},
  doi = {10.1162/coli_a_00481}
}

@misc{suttacentral-nd,
  author = {{SuttaCentral}},
  title = {{SuttaCentral}},
  year = {n.d.},
  note = {Accessed April 30, 2026},
  url = {https://suttacentral.net/index.html?lang=en}
}

@misc{thanissaro-nd-suttas,
  author = {{Thanissaro Bhikkhu}},
  title = {Suttas},
  year = {n.d.},
  note = {Accessed April 30, 2026},
  url = {https://www.dhammatalks.org/suttas/}
}

@inproceedings{thompson-post-2020-paraphrasing,
  author = {Thompson, Brian and Post, Matt},
  title = {Automatic Machine Translation Evaluation in Many Languages via Zero-Shot Paraphrasing},
  booktitle = {Proceedings of the 2020 Conference on Empirical Methods in Natural Language Processing},
  pages = {90--121},
  year = {2020},
  doi = {10.18653/v1/2020.emnlp-main.8}
}

@inproceedings{vaswani-etal-2017-attention,
  author = {Vaswani, Ashish and Shazeer, Noam and Parmar, Niki and others},
  title = {Attention Is All You Need},
  booktitle = {Advances in Neural Information Processing Systems 30},
  pages = {5998--6008},
  year = {2017},
  note = {arXiv:1706.03762}
}

@book{van-buitenen-1981-bhagavadgita,
  author = {van Buitenen, J. A. B.},
  title = {The Bhagavadg{\=i}t{\=a} in the Mah{\=a}bh{\=a}rata: Text and Translation},
  publisher = {University of Chicago Press},
  year = {1981}
}

@book{walshe-1995-long,
  author = {Walshe, Maurice},
  title = {The Long Discourses of the Buddha: A Translation of the D{\=i}gha Nik{\=a}ya},
  publisher = {Wisdom Publications},
  year = {1995}
}

@misc{wu-etal-2024-multiple,
  author = {Wu, Si and Wieting, John and Smith, David A.},
  title = {Multiple References with Meaningful Variations Improve Literary Machine Translation},
  year = {2024},
  note = {arXiv preprint, arXiv:2412.18707}
}

@inproceedings{xu-etal-2024-paradigm,
  author = {Xu, Haoran and Kim, Young Jin and Sharaf, Amr and Awadalla, Hany Hassan},
  title = {A Paradigm Shift in Machine Translation: Boosting Translation Performance of Large Language Models},
  booktitle = {The Twelfth International Conference on Learning Representations},
  year = {2024},
  note = {arXiv:2309.11674}
}

@inproceedings{zhang-etal-2020-bertscore,
  author = {Zhang, Tianyi and Kishore, Varsha and Wu, Felix and Weinberger, Kilian Q. and Artzi, Yoav},
  title = {{BERTScore}: Evaluating Text Generation with {BERT}},
  booktitle = {The Eighth International Conference on Learning Representations},
  year = {2020},
  note = {arXiv:1904.09675}
}

@misc{zhang-etal-2025-qwen3,
  author = {Zhang, Yanzhao and Li, Mingxin and Long, Dingkun and others},
  title = {{Qwen3} Embedding: Advancing Text Embedding and Reranking Through Foundation Models},
  year = {2025},
  note = {arXiv preprint, arXiv:2506.05176. Model used: Qwen3-Embedding-8B},
  url = {https://huggingface.co/Qwen/Qwen3-Embedding-8B}
}

@inproceedings{zhu-etal-2024-multilingual,
  author = {Zhu, Wenhao and Liu, Hongyi and Dong, Qingxiu and others},
  title = {Multilingual Machine Translation with Large Language Models: Empirical Results and Analysis},
  booktitle = {Findings of the Association for Computational Linguistics: NAACL 2024},
  pages = {2765--2781},
  year = {2024},
  doi = {10.18653/v1/2024.findings-naacl.176}
}

\clearpage
\onecolumn
\appendix
\raggedright
\section{Translation Prompt}

The following system prompt was used for all model translations via \texttt{scripts/benchmark\_translate.py}. Passages were batched by token count (default 3,000 tokens per batch) and sent as a JSON object mapping passage identifiers to Pali text. The model was instructed to return a JSON object with the same keys and English translations as values.

\begin{Verbatim}[fontsize=\footnotesize,commandchars=\\\{\}]
Translate the following Pali passages into English.
Return a JSON object with the same keys and the English translations as values.
Do not add text that is not present in the input.

IMPORTANT: For valid JSON, never use double quotes inside translation values.
- Use single quotes for dialogue: 'Hello,' he said.
- Use backticks for nested quotes: 'He said `hello` to me.'

Example:
Input: \{"mn1:5": "Pathav\=i pathav\=iti sa\~nj\=an\=ati.",
        "mn1:6": "\=Apo \=apoti sa\~nj\=an\=ati."\}
Output: \{"mn1:5": "They perceive earth as earth.",
         "mn1:6": "They perceive water as water."\}

Output ONLY the JSON object, no other text.
\end{Verbatim}

\section{Alignment Prompt}

The following system prompt was used by \texttt{pipeline/align.py} to align Thanissaro and Bodhi translations to SuttaCentral segment identifiers. It is reproduced verbatim, preserving the prompt's original spelling.

\begin{Verbatim}[fontsize=\footnotesize,commandchars=\\\{\}]
ROLE
You are an alignment engine. Your task is to extract English text
from an existing translation to match specific P\=ali segment IDs.

INPUT
For each segment ID you are given:
- the P\=ali text (semantic anchor - this determines what content
  belongs to the segment)
- Sujato's English translation (segmentation reference only)

You are also given the FULL English translation by the target translator.

GOAL
For every segment ID, extract the corresponding English text from
 the translator that expresses the same meaning as the P\=ali.

IMPORTANT RULES

1) P\=ali is the authority.
   Use the P\=ali text to decide what content belongs to each segment.

2) Sujato is only a guide.
   His segmentation helps you understand boundaries, but the target
   translator may split/merge differently.

3) MATCH THE PALI STRUCTURE.
   The OUTPUT must mirror the structure of the PALI segment:

   a) If the Pali contains "..." or is abbreviated:
      -> Extract ONLY the corresponding term(s) from the translator
      -> Do NOT expand to the full sentence
      -> Example: Pali "vi\~n\~n\=ata\d{m} \ldots{}"
         + Sujato "the known \ldots{}"
        -> Output just "the cognized" (not the full paragraph)

   b) If the Pali is fully spelled out (no ellipsis):
      -> Extract the complete matching text from the translator
      -> If the translator abbreviates with "..." but Pali is full,
        expand by finding the earlier template and substituting the terms

4) Minimal faithful extraction.
   Extract the smallest text that expresses the P\=ali meaning.
   Prefer contiguous substrings from the original.

5) Respect text order.
   Process the translation in order. Don't reuse non-repetitive text.

6) Filter noise.
   Discard footnotes, section headers (unless part of translation),
   editor notes, bracketed references.

7) Null policy.
   Output null ONLY if you genuinely cannot find matching content.
   Remember: abbreviated Pali -> short output (just the term).

OUTPUT FORMAT (STRICT JSON)
Return valid JSON with exactly the same keys as the input,
in the same order.
Each key maps to either a string (extracted text) or null.

Example:
\{
  "mn1:3.1": "Here, monks, an untaught ordinary person...",
  "mn1:3.2": "He perceives earth as earth.",
  "mn1:3.3": null
\}

Do NOT include P\=ali, Sujato, or any extra fields.
Do NOT add commentary or explanation.
JSON only.
\end{Verbatim}

\end{document}